\RequirePackage{fix-cm}
\documentclass[twocolumn,compsoc]{CVM}
\pdfoutput=1
\usepackage{times}
\usepackage{epsfig}
\usepackage{graphicx}
\usepackage{amsmath}
\usepackage{amssymb}
\usepackage{url}
\usepackage{xcolor}
\usepackage{overpic}
\usepackage{multirow}
\usepackage[switch,pagewise]{lineno} 

\definecolor{red}{rgb}{0.8,0.2,0.2}

\definecolor{green}{rgb}{0.2,0.8,0.2}

\def\ManuscriptInfo{Manuscript received: 20xx-xx-xx; accepted: 20xx-xx-xx.}

\def\PaperTitle{Recurrent 3D attentional networks for end-to-end  active object recognition}

\def\AuthorNames{Min Liu$^{1,2}$, Yifei Shi$^1$, Lintao Zheng$^1$, Kai Xu$^1$\rm(\rm{\Letter})\textbf{,} \textbf{Hui Huang}$^3$\textbf{,} \textbf{and Dinesh Manocha}$^2$}

\def\AuthorAdress{
$1\quad $ & School of Computer, National University of Defense Technology, Changsha, 410073, China. E-mail: M. Liu, gfsliumin@gmail.com; Y. Shi, jerrysyf@gmail.com; L. Zheng, lintaozheng1991@gmail.com; K. Xu, kevin.kai.xu@gmail.com\rm(\rm{\Letter}).\\
$2\quad $ & Department of Computer Science and Electrical \& Computer Engineering, University of Maryland, College Park, 20742, USA. E-mail: D. Manocha, dm@cs.umd.edu.\\
$3\quad $ & Visual Computing Research Center, Shenzhen University, Shenzhen, 518060, China. E-mail: huihuang@szu.edu.cn.\\
}

\def\firstheader{}

\headevenname{{M. Liu, Y. Shi, L. Zheng, \etal} }

\begin{document}

\MakePageStyle

\MakeAbstract{Active vision is inherently attention-driven: The agent actively selects views to attend in order to fast achieve the vision task while improving its internal representation of the scene being observed. Inspired by the recent success of attention-based models in 2D vision tasks based on single RGB images, we propose to address the multi-view depth-based active object recognition using attention mechanism, through developing an end-to-end recurrent 3D attentional network. The architecture takes advantage of a recurrent neural network (RNN) to store and update an internal representation. Our model, trained with 3D shape datasets, is able to iteratively attend to the best views targeting an object of interest for recognizing it. 
To realize 3D view selection, we derive a 3D spatial transformer network which is differentiable for training with backpropagation, achieving much faster convergence than the reinforcement learning employed by most existing attention-based models. Experiments show that our method, with only depth input, achieves state-of-the-art next-best-view performance in time efficiency and recognition accuracy.
}

\MakeKeywords{active object recognition; recurrent neural network; next-best-view; 3D attention}

\section{Introduction}
\label{sec:intro}

Active object recognition plays a central role in robot-operated autonomous scene understanding
and object manipulation. The problem involves online planning the views of the visual sensor
of a robot to maximally increase the accuracy and confidence of object recognition, which is also referred to as the next-best-view (NBV) problem for active object recognition.
Recently, 3D object recognition has gained much boosting thanks to the fast development of 3D sensing techniques (e.g., depth cameras) and the proliferation of 3D shape repositories. Our work also adopts the 3D geometric data-driven approach, under the setting of 2.5D depth acquisition.

For view selection, most existing works lie in the paradigm of information theoretic view evaluation (e.g.~\cite{denzler2002,huber2012}).
For example, from a set of candidates, the view maximizing the mutual information between observations and object classes
is selected. Such methods often present two issues.
First, to estimate the mutual information, unobserved views must be sampled and
the corresponding data must be synthesized from a learned generative model,
making the view estimation inefficient~\cite{Wu15}.
Second, the object recognition model is typically learned independent on the view planner, although the two
are really coupled in an active recognition system~\cite{jayaraman2016}.

Some works formulate active recognition as a reinforcement learning problem,
to learn a viewing policy under various observations.
Especially, a few recent works attempted end-to-end reinforcement learning
based on recurrent neural networks~\cite{jayaraman2016,xu2016,veram2018}.
Applying a learned policy is apparently much more efficient than sampling from a generative model.
However, these models are known to be hard to train,
with difficult parameter tuning and relatively long training time~\cite{Mnih2014}.
Moreover, the success of these methods highly depends on the hand-designed reward functions.

\begin{figure*}[t!] \centering
    \begin{overpic}[width=0.75\linewidth,tics=5]{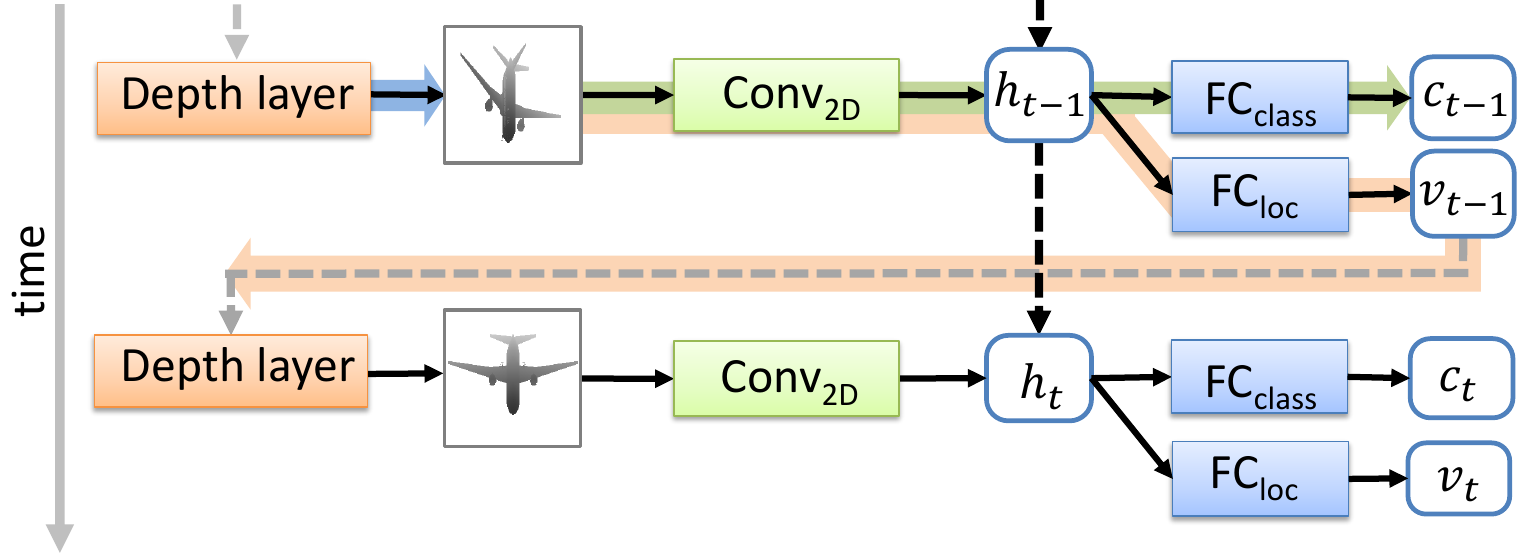}
    \end{overpic}
    \caption{Our recurrent attentional model. The dashed arrows indicate information flow across time steps while the solid ones represent that within a time step. The bold arrows underline the data flows of the three subnetworks, i.e., depth layer (DL, blue), 3D spatial transformer networks (3D-STN, orange) and shape classifier (SC, green).
    }
    \label{fig:model}
    \vspace{-12pt}
\end{figure*}

The recent development of attention-based deep models has led to significant success
in 2D vision tasks based on RGB images~\cite{Mnih2014,Xu2015Cap}.
Attention-based models achieve both efficiency and accuracy by focusing the processing
only on the most informative parts of the input with respect to a given task.
Information gained from different fixations is integrated into an internal representation,
to approach the desired goal and guide future attention.
Such mechanism, being both goal-directed and stimulus-driven~\cite{Corbetta2002},
fits well to the problem setting of active recognition,
accommodating object recognition and guided acquisition in a unified optimization.

However, the popular formulation of attention model based on recurrent neural networks~\cite{Mnih2014}
suffers from the problem of indifferentiable recognition loss over the attentional locations, making
the network optimization by backpropagation infeasible.
To make it learnable, the training is often turned into
a partially observable Markov decision process (POMDP), which comes back to reinforcement learning.
The recently introduced differentiable Spatial Transformer Networks (STN) can be
used to actively predict image locations for 2D object detection and recognition~\cite{jaderberg2015spatial}.
Motivated by this, we opt to use STN units as our localization networks.

However, extending the standard STN to predict views in 3D space
while keeping its differentiability is non-trivial.
To facilitate the backpropagation of loss gradient from a 2.5D depth image to 3D viewing parameters,
we propose to parameterize the depth value at each pixel $(x,y)$ in a depth image over the parameters of the corresponding
view $(\theta,\varphi)$:
$d(x,y)=f(\theta,\varphi)$, through a ray casting based depth computation at each pixel.
Our attention model produces efficient view planning and robust object recognition, as demonstrated by experimental evaluations. Our work contains two main technical contributions:
\begin{itemize}
\item A 3D attentional architecture that integrates RNN and STN for simultaneous object recognition and next-best-view (NBV) selection.
\item A differentiable extension of STN for view selection in 3D space, leading to an end-to-end attentional network which can be trained efficiently.

\end{itemize}


\section{Related work}
\label{sec:related}

Active object recognition has a rich literature in robotics, vision and graphics (surveys available from e.g.~\cite{scott2003,roy2004active}). We provide a brief review for 3D object recognition,
especially those active methods (categorized into information
theoretic and policy learning approaches). We then discuss some
recent attempts on end-to-end learning for NBV selection.

\paragraph*{3D object recognition.}
One of the most popular methods for 3D object recognition is
directly deploying deep learning on point sets~\cite{qi2016pointnet,qi2017pointnetplusplus}, one shortcoming for these works is that point features are treated independently. SCORES\cite{zhu2018scores} uses a recursive neural network for shape composition, and the representation of shape composition can be used for 3D object recognition. Based on pioneering work, Attentional ShapeContextNet~\cite{Xie_2018_CVPR}, which connects shape context with convolutional neural networks (CNN), is able to represent the local and global shape information, and gain competitive results on benchmark datasets. Inspired by image classification using CNN, view-based methods for 3D object recognition have performed best so far. Multi-view CNN~\cite{Wu15} renders a 3D shape to gray images from different views, uses CNN to extract features for each rendered image, and aggregates features from all rendered images with max pooling. A hierarchical view-group-shape architecture is proposed~\cite{Feng_2018_CVPR} aiming to treat each view discriminatively, while all views are treated equally in~\cite{Wu15}. Triangle lasso~\cite{zhao2018triangle} can be used for simultaneous clustering and optimization in graph datasets, and is potential for views aggregation.  Impressive improvement is gained in~\cite{Feng_2018_CVPR}, but it still need evenly sample several views before testing, which means all views are fixed when testing. This kind of method is not suitable for some scenarios, such as robot-operated NBV problem, which always tries to achieve the highest accuracy with as few as possible views.

\paragraph*{Information theoretic approaches.}
Information theoretic formulation represents a standard approach to active vision problems.
The basic idea is to quantify the information gain of each view by measuring
the mutual information between observations and object classes~\cite{denzler2002},
the entropy reduction of object hypotheses~\cite{borotschnig2000},
or the decrease of belief uncertainty about the object that generated the observations~\cite{callari2001}.
The optimal views are those which are expected to receive the maximal information gain.
The estimation of information gain usually involves learning a generative object model
(likelihood or belief state) so that the posterior distribution of object class
under different views can be estimated.
Different methods have been utilized in estimating information gain, such as
Monte Carlo sampling~\cite{denzler2002},
Gaussian Process Regression~\cite{huber2012} and reinforcement learning~\cite{arbel2001entropy}.

\paragraph*{Policy learning approaches.}
Another line of research seeks to learn viewing policies.
The problem is often viewed as a stochastic optimal control one and cast as
a partially-observable Markov decision process.
In~\cite{paletta2000active}, reinforcement learning is utilized to offline learn an approximate policy
that maps a sequence of observations to a discriminative viewpoint.
Kurniawati et al. employ a point-based approximate solver to obtain a non-greedy policy offline~\cite{kurniawati2008sarsop}.
In contrast to offline learning, Lauri et al. attempted to apply Monte Carlo tree search (MCTS)
to obtain online active hypothesis testing policy for view selection~\cite{lauri2015active}.
Our method learns and compiles viewing policies into the hidden layers of RNN, leading to a high-capacity view planning model.

\paragraph*{End-to-end learning approaches.}
The recent fast development of deep learning models has aroused the interest of end-to-end learning
of active vision policies~\cite{levine2016end}.
Malmir et al. use deep Q-learning to find the optimal policy for view selection from raw images~\cite{malmir2016deep}.
Our method shares similarities with the recent work of Wu et al.~\cite{Wu15}
in taking 2.5D depth images as input and 3D shapes as training data.
They adopt a volumetric representation of 3D shapes and train a Convolutional Deep Belief Network (CDBN)
to model the joint distribution over volume occupancy and shape category.
By sampling from the distribution, shape completion can be performed based on observed depth images,
over which virtual scanning is conducted to estimate the information gain of a view.
Different from their method, our attention model is trained offline hence no online sampling is required,
making it efficient for online active recognition.
The works of Jayaraman and Grauman~\cite{jayaraman2016}, Xu et al.~\cite{xu2016} and Chen et al.~\cite{veram2018} are the most related to ours. Compared to MV-RNN~\cite{xu2016}, VERAM~\cite{veram2018} explicitly integrates view confidence and view location constrains into reward function, and deploy some strategies to do view-enhancement. In these methods, the recognition and control modules are jointly optimized based on reinforcement learning. We employ the spatial transformer units~\cite{jaderberg2015spatial} as our locator networks to obtain a fully differentiable network.

\paragraph*{Spatial Transformer Networks.}
Spatial transformer is a differentiable sampling-based
network, which gives neural networks the ability
to actively transform the input data, without any spatial supervision in training.
Generally, spatial transformer is composed of a localization network and a generator.
STN achieves spatial attention by first passing the input image into a localization network
which regresses the transformation,
and then generating a transformed image by the generator.
The transformed image is deemed to be easier to recognize or classify, thus better approaching the desired task.

STN fits well to our problem setting.
Due to its differentiability, it enables end-to-end training with backpropagation, making the network easier to learn. 
It is relatively straightforward to employ it for object localization in the image of a given view.
However, when using it to predict views in 3D, we face the problem of indifferentiable pixel depth values
over viewing parameters, which is addressed by our work.

\section{Approach}
\label{sec:method}
\begin{figure}[t!] \centering
    \begin{overpic}[width=0.8\linewidth,tics=5]{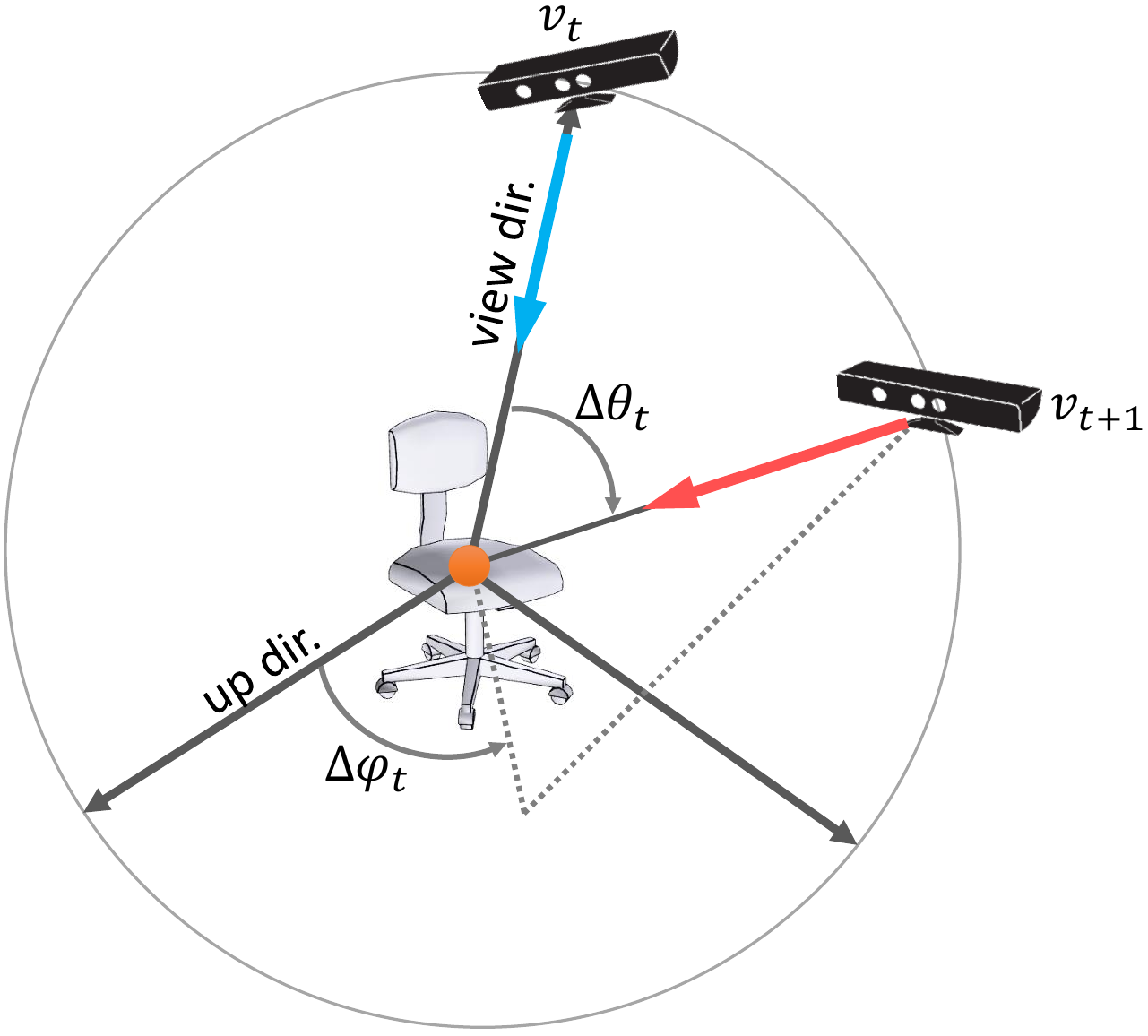}
    \end{overpic}
    \caption{The viewing sphere around the object being recognized. The view of the next time step, $v_{t+1}$, is parameterized in the local spherical coordinate system w.r.t. the current view $v_{t}$.
    }
    \label{fig:sphere}
    \vspace{-12pt}
\end{figure}

We first provide an architectural overview of our recurrent attentional model, followed by detailed introduction of the individual parts. We also provide details about loss function, model training and inference.

\subsection{Architecture Overview}
Fig.~\ref{fig:model} shows the architecture of our recurrent attentional model. The main body of the model is a Recurrent Neural Network (RNN) for modeling the sequential dependencies among consecutive views.
In our settings, the 3D shape is located in the center of a sphere with fixed radius (Fig.~\ref{fig:sphere}), and at each time step, the model first takes a view (parameterized in the local sphere coordinate system) as input, generates depth images using ray casting and extracts features of depth images. Then the model amalgamates information of past views, makes a prediction of the categorical label and produces the increment of current view for future observation.
To achieve that, we embed three parts into the RNN:
a depth layer (DL) for generating depth images of objects,
a 3D spatial transformer network (3D-STN) for regressing the increment of current view for the shape, and a shape classifier (SC) for depth-based object recognition.

Our model works as follows.
Given a current view of an object, it first uses ray casting algorithm to generate a depth image, then the depth image is fed into a stack of convolutional layers (Conv$_{\text{2D}}$) for feature extraction.
The extracted features are aggregated with those extracted from the past views, with the help of the RNN hidden layers.
The aggregated features are then used for classification and predicting the increment of current view, with the fully connected layers FC$_{\text{class}}$ and FC$_{\text{loc}}$, respectively.
With the predicted increment of current view, a next-best-view (${v_{t+1}} = {v_{t}} + \Delta v_{t}$) can be obtained. Using $v_{t+1}$, a new depth image can be generated, which serves as the input to the next time step.

As shown in Fig.~\ref{fig:model}, DL is a single layer subnetwork.
3D-STN encompasses the convolutional layers Conv$_{\text{2D}}$ and the fully connected layers FC$_{\text{loc}}$.
SC is composed of Conv$_{\text{2D}}$ and FC$_{\text{class}}$, which is a standard Convolutional Neural Network (CNN) classifier. Moreover, the convolutional layers are shared by the SC and the 3D-STN.

In order to make our attentional network learnable,
we require the generated depth image differentiable against the viewing parameters.  A basic assumption behind the parameterization of depth values w.r.t. viewing parameters and the intersection point is that the intersection point does not change when the view change is small, thus the first-order derivative over viewing parameters can be approximated by keeping the point fixed. Details in \ref{backpropagation}.

\subsection{Depth Layer for Depth Images Generation}
Depth layer (DL) is a critical part of our model, including depth images generation and loss backpropagation. With loss backpropagation, this layer allows us to build up an end-to-end training fashion deep neural network. During forward propagation, we use ray casting to generate depth images, and record every hit point position on the surface of shapes into a table, which will be used in backpropagation. During backpropagation, we fill the gap between depth images loss gradient and camera views loss gradient.

\begin{figure}[t!] \centering
    \begin{overpic}[width=0.8\linewidth,tics=5]{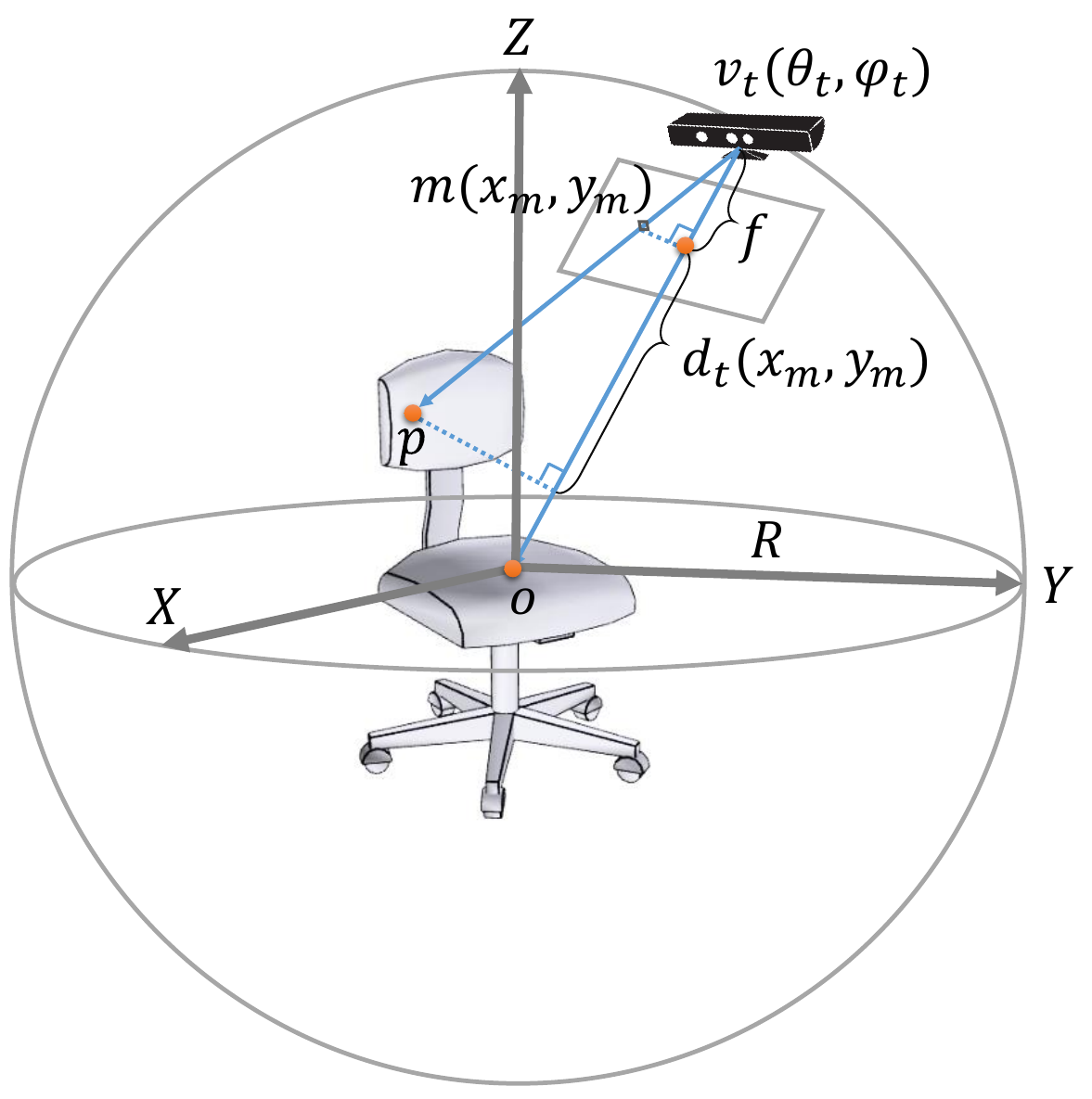}
    \end{overpic}
    \caption{Given a view $v_t=(\theta_t,\varphi_t)$, the depth value of a pixel, $d_t(x_m,y_m)$, in the corresponding depth
    image is parameterized against the view parameters and the position of the intersection point, $p$, computed by ray-casting.
    }
    \label{fig:distance}
    \vspace{-12pt}
\end{figure}

\paragraph*{Ray casting.}
Ray casting is to solve the general problem of determining the hit points of shape intersected by a ray. As illustrated in Fig.~\ref{fig:distance}, a view is represented by ($R,\theta_t,\varphi_t$), where $R$, $\theta_t$ and $\varphi_t$ are the radius distance to the shape center, polar and azimuthal angle, respectively. In this spherical coordinate system,  $R$ is a constant, and the view  direction points to the shape's center, which is also the origin of spherical coordinate system. The $(R ,\theta_t ,\varphi_t)$ can be easily transformed into Cartesian coordinate $(X_v ,Y_v ,Z_v)$ by using Eq.~(\ref{eq:coor}), where ($\theta_t,\varphi_t$) can be obtained by Eq.~(\ref{eq:incre}).
\begin{equation}
\label{eq:coor}
\left\{ \begin{array}{l}
X_v = R \sin \theta_t cos\varphi_t \\
Y_v = R \sin \theta_t sin\varphi_t \\
Z_v = R \cos \theta_t
\end{array} \right.
\end{equation}

\begin{equation}
\label{eq:incre}
\left\{ \begin{array}{l}
{\theta _t} = {\theta _{t - 1}} + \Delta {\theta _{t-1}}\\
{\varphi _t} = {\varphi _{t - 1}} + \Delta {\varphi _{t-1}}
\end{array} \right..
\end{equation}

As illustrated in Fig.~\ref{fig:distance}, a projection plane lies between the viewpoint $v_t$ and the shape. $m$ is the intersection point of view line and projection plane. For ray casting, we need two points (or one point along with a direction). The position of $v_t$ can be directly computed by using Eq.~(\ref{eq:coor}). However, for the position of $m$, it is not straightforward. To make things simpler, we have a specific setting as follows: When $\theta_t = 0$, the position of $v_t$ is $(0,0,R)$, the up direction of camera is set to be parallel to $X$-axis of spherical coordinate system, and the origin of pixel coordinate is at $(0,0,R-f)$, where $f$ is the focal length of camera. From projection coordinate to pixel coordinate, the camera intrinsic matrix should be applied. If we have each pixel position $(u_m,v_m)$ in pixel coordinate, the projection coordinate $(x_m,y_m)$ can be calculated by Eq.~(\ref{eq:inv_intr}).
\begin{equation}
\label{eq:inv_intr}
\left[ \begin{array}{l}
{x_m}\\
{y_m}\\
1
\end{array} \right] = \left[ {\begin{array}{*{20}{c}}
{dx}&0&{ - {u_0}dx}\\
0&{dy}&{ - {v_0}dy}\\
0&0&1
\end{array}} \right]\left[ \begin{array}{l}
{u_m}\\
{v_m}\\
1
\end{array} \right],
\end{equation}
where $dx,dy$ is single pixel length along $x$-axis and $y$-axis of projection coordinate, $u_0,v_0$ represent the principal point, which would be ideally in the center of depth images, and skew coefficient between $x$-axis and $y$-axis is set to 0. Since we have a pre-defined up direction of the camera, we can compute the world coordinate position of each pixel in projection plane. When $v_t$ is located in ($R,\theta_t,\varphi_t$), we just apply 2 rotation matrix to get the new Cartesian coordinate position of each pixel, the first rotation is $\theta_t$ around $Y$-axis, and the second rotation is $\varphi_t$ around $Z$-axis.

For each pixel in projection plane, we can form a ray from $v_t$ to $m$, and the extention of ray will (or not) hit the shape surface. Using ray casting algorithm, we can get the hit point ($p$) position $({X_p},{Y_p},{Z_p})$ in the shape surface (if hit) and the hit distance ($D_{vp}$) between the start point ($v_t$) and hit point ($p$). As shown in Fig.~\ref{fig:distance}, if a ray hits the shape, the depth value of related pixel is represented by $d_t(x_m,y_m)$. The distance between $v_t$ and $p$ is calculated in Eq.~(\ref{eq:act_dist}). And according to similar triangle theorem, we have Eq.~(\ref{eq:distance}). We use a table to record all hit points and their related pixels.

\begin{equation}
\label{eq:act_dist}
{D_{vp}} = \sqrt {{{({X_v} - {X_p})}^2} + {{({Y_v} - {Y_p})}^2} + {{({Z_v} - {Z_p})}^2}}
\end{equation}
\begin{equation}
\label{eq:distance}
{d_t}({x_m},{y_m}) = \frac{f}{{\sqrt {x_m^2 + y_m^2 + {f^2}} }} \cdot {D_{vp}} - f.
\end{equation}

\paragraph*{Backpropagation.}
\label{backpropagation}
A backpropagation through the depth layer computes loss gradients of input ($\Delta {\theta _{t-1}},\Delta {\varphi _{t-1}}$), given loss gradients of output (depth image). During backpropagation, each pixel will get a ${\partial loss}/{\partial d_t}$. We obtain the depth value $d_t$, and the hit point position $({X_p},{Y_p},{Z_p})$ (which can be directly retrieved from the recorded table), so ${{\partial loss}}/{{\partial \Delta{\theta_{t-1} }}}$ and ${{\partial loss}}/{{\partial \Delta{\varphi_{t-1} }}}$ in Eq.~(\ref{eq6}) can be calculated along with Eq.~(\ref{eq:coor}),~(\ref{eq:incre}),~(\ref{eq:act_dist}),~(\ref{eq:distance}) and~(\ref{eq5}).
\begin{equation}
\label{eq5}
\left\{ \begin{array}{l}
\frac{{\partial {d_t}}}{{\partial \Delta {\theta _{t - 1}}}} = \frac{{\partial {d_t}}}{{\partial {D_{vp}}}} \times \frac{{\partial {D_{vp}}}}{{\partial {X_v}}} \times \frac{{\partial {X_v}}}{{\partial {\theta _t}}} \times \frac{{\partial {\theta _t}}}{{\partial \Delta {\theta _{t - 1}}}} + \\
\ \ \quad\quad\quad\quad{\rm{             }}\frac{{\partial {d_t}}}{{\partial {D_{vp}}}} \times \frac{{\partial {D_{vp}}}}{{\partial {Y_v}}} \times \frac{{\partial {Y_v}}}{{\partial {\theta _t}}} \times \frac{{\partial {\theta _t}}}{{\partial \Delta {\theta _{t - 1}}}} + \\
\ \ \quad\quad\quad\quad{\rm{             }}\frac{{\partial {d_t}}}{{\partial {D_{vp}}}} \times \frac{{\partial {D_{vp}}}}{{\partial {Z_v}}} \times \frac{{\partial {Z_v}}}{{\partial {\theta _t}}} \times \frac{{\partial {\theta _t}}}{{\partial \Delta {\theta _{t - 1}}}}\\
\frac{{\partial {d_t}}}{{\partial \Delta {\varphi _{t - 1}}}} = \frac{{\partial {d_t}}}{{\partial {D_{vp}}}} \times \frac{{\partial {D_{vp}}}}{{\partial {X_v}}} \times \frac{{\partial {X_v}}}{{\partial {\varphi _t}}} \times \frac{{\partial {\varphi _t}}}{{\partial \Delta {\varphi _{t - 1}}}} + \\
\ \ \quad\quad\quad\quad{\rm{              }}\frac{{\partial {d_t}}}{{\partial {D_{vp}}}} \times \frac{{\partial {D_{vp}}}}{{\partial {Y_v}}} \times \frac{{\partial {Y_v}}}{{\partial {\varphi _t}}} \times \frac{{\partial {\varphi _t}}}{{\partial \Delta {\varphi _{t - 1}}}} + \\
\ \ \quad\quad\quad\quad{\rm{              }}\frac{{\partial {d_t}}}{{\partial {D_{vp}}}} \times \frac{{\partial {D_{vp}}}}{{\partial {Z_v}}} \times \frac{{\partial {Z_v}}}{{\partial {\varphi _t}}} \times \frac{{\partial {\varphi _t}}}{{\partial \Delta {\varphi _{t - 1}}}}
\end{array} \right.
\end{equation}

\begin{equation}
\label{eq6}
\left\{ \begin{array}{l}
\frac{{\partial loss}}{{\partial \Delta{\theta_{t-1} } }} = \frac{{\partial loss}}{{\partial d_t}} \times \frac{{\partial d_t}}{{\partial \Delta{\theta_{t-1} }}}\\\\
\frac{{\partial loss}}{{\partial \Delta{\varphi_{t-1} }}} = \frac{{\partial loss}}{{\partial d_t}} \times \frac{{\partial d_t}}{{\partial \Delta{\varphi_{t-1} }}}
\end{array} \right.,
\end{equation}
each pixel of depth image will get a ${{\partial loss}}/{{\partial \Delta{\theta_{t-1} }}}$ and ${{\partial loss}}/{{\partial \Delta{\varphi_{t-1} }}}$, we just average all the ${{\partial loss}}/{{\partial \Delta{\theta_{t-1} }}}$ and ${{\partial loss}}/{{\partial \Delta{\varphi_{t-1} }}}$ to get the final loss gradients of $(\Delta{\theta_{t-1}} ,\Delta{\varphi_{t-1}} )$.

\subsection{3D-STN for View Selection}
Given an input depth image, the goal of 3D-STN is to extract image features, regress the 3D viewing parameters of the  increment of current view ($\Delta {\theta _t},\Delta {\varphi _t}$) with respect to the recognition task.
3D-STN is comprised of two subnetworks: Conv$_{\text{2D}}$ and  FC$_{\text{loc}}$. During forward passing, Conv$_{\text{2D}}$ takes the depth image $d_t$ as input and extracts features. With the features, FC$_{\text{loc}}$outputs the increment of the current view. The viewing parameter is parameterized in the local spherical coordinate system w.r.t. the current view $v_t$ ($\theta _t,\varphi _t$) (Fig.~\ref{fig:sphere}),
and represented as a tuple ($\Delta {\theta _t},\Delta {\varphi _t}$).
Note that the viewing parameters does not include radius ($R$), since $R$ is set to be a constant.

Specifically, the convolutional network Conv$_{\text{2D}}$ first extracts features from the depth image
output by depth layer:
\begin{equation}
e_t = f_{\text{Conv}}^{\text{2D}}(d_t, W_{\text{Conv}}^{\text{2D}}),
\label{eq:3dstn1}
\end{equation}
where $W_{\text{Conv}}^{\text{2D}}$ are the weights of Conv$_{\text{2D}}$.
These features are amalgamated with those of past views stored in the RNN hidden layer $h_{t-1}$:
\begin{equation}
h_{t} = g(W_{ih}e_t + W_{hh}h_{t-1}),
\label{eq:rnn}
\end{equation}
where $g(\cdot)$ is a nonlinear activation function.
$W_{ih}$ and $W_{hh}$ are the weights for input-to-hidden and hidden-to-hidden connections, respectively.
The aggregated features in $h_t$ is then used to regress the increment of current view:
\begin{equation}
(\Delta{\theta_{t}},\Delta{\varphi_{t})} = f_{\text{FC}}^{\text{loc}}(h_t, W_{\text{FC}}^{\text{loc}}),
\label{eq:3dstn2}
\end{equation}
where $W_{\text{FC}}^{\text{loc}}$ are the weights of FC$_{\text{loc}}$.
The viewing parameters are then used by the depth layer to generate a new depth image with the object during training:
\begin{equation}
d_{t+1} = f_{\text{DL}}(\theta_{t}+\Delta{\theta_{t}},\varphi_{t}+\Delta{\varphi_{t})}).
\label{eq:3dstn3}
\end{equation}

%

\subsection{Shape Classifier for Object Recognition}
The depth image output by the depth layer at each time step is passed into a shape classifier (SC, Conv$_{\text{2D}}$ + FC$_{\text{class}}$) for class label prediction.
Note that the classification is also based on the aggregated features
of both current and past views:
\begin{equation}
c_t = f_{\text{FC}}^{\text{class}}(h_t, W_{FC}^{\text{class}}),
\label{eq:class}
\end{equation}
where $W_{\text{FC}}^{\text{class}}$ are the weights of FC$_{\text{class}}$.

\begin{figure}[t!] \centering
    \begin{overpic}[width=\linewidth,tics=5]{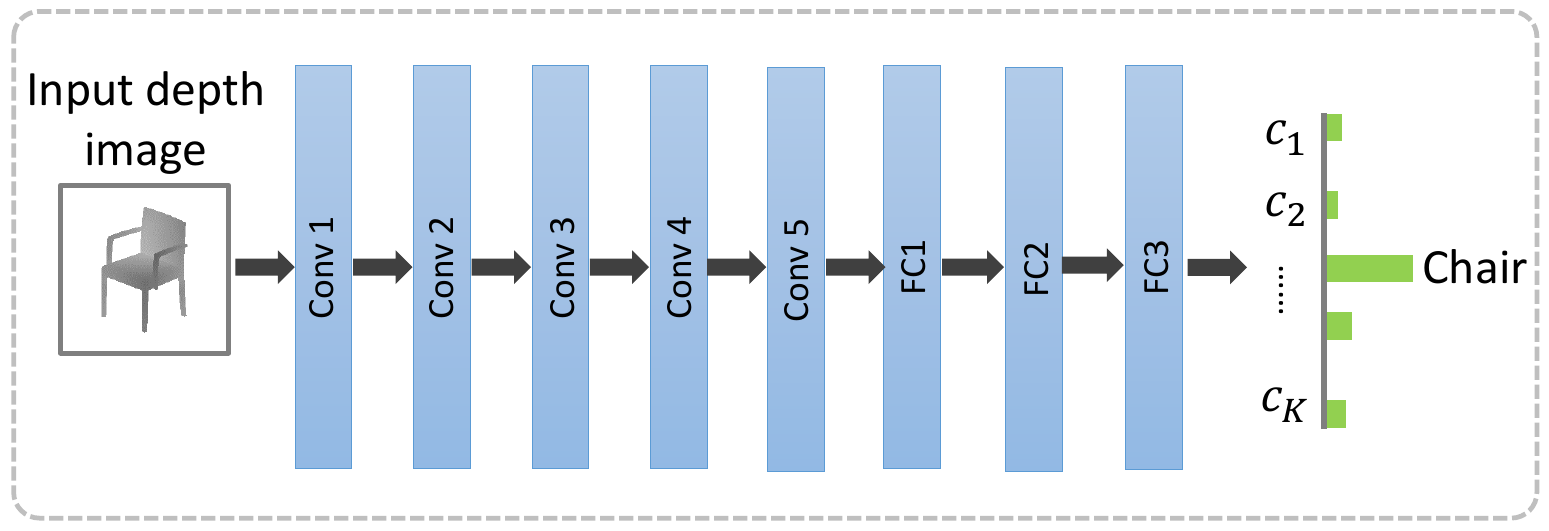}
    \end{overpic}
    \caption{The architecture of our shape classifier (SC), which is borrowed from AlexNet~\cite{krizhevsky2012}. SC takes depth images as input, and outputs classification results.}
    \label{fig:2dstn}
    \vspace{-12pt}
\end{figure}

\subsection{Loss function}
We employ cross-entropy loss to train our model. Cross-entropy loss measures the performance of a classification model whose output is a probability value between 0 and 1. Our loss function is:
\begin{equation}
L =  - \sum\limits_{c = 1}^k {{y_{o,c}}\log ({p_{o,c}})},
\label{eq:loss}
\end{equation}
where $k$ is the number of classes, $y$ is a binary indicator indicating whether class label $c$ is the correct classification for current observation $o$, and $p$ is the predicted probability observation $o$ is of class $c$.

\subsection{Training and Inference}
To make the training more efficient, we propose to decompose our training into two parts and tune their parameters separately.:
1) pre-training SC;
2) joint training 3D-STN, SC and RNN.
The first part is trained by virtually recognizing 3D shapes in the training dataset, using generated depth images, we hope the Conv$_{\text{2D}}$ of SC have a good ability to extract features for depth images. For each shape, we randomly select dozens of views to generate depth images, and feed them to a pre-trained CNN classifier (we use AlexNet~\cite{krizhevsky2012}) as shown in Fig.~\ref{fig:2dstn}, in order to train it with respect to an image classification task.

For training 3D-STN, we evenly select 50 views as the initial views, and from each initial view, we start the virtual recognizing for an episode of 10 time steps. In each step, the network takes images as input, and outputs both class label and the increment of the current view. The network is trained the cross entropy loss in classification. The training leverage parameters of the pre-trained SC, and tunes the parameters of the 3D-STN, the RNN and SC simultaneously, using backpropagation through time (BPTT)~\cite{Mozer1989}. The number of initial views is a trade-off between networks performance and computation density. With more initial views to explore, networks will obtain better performance, but the training time will be greatly increased. We choose 50 as the number of initial views such that our networks can obtain a good performance and the training time is affordable.

At inference time, given a object, a depth image is generated using ray casting algorithm with a random initial view from our selected 50 views. The generated depth image is then passed into our attentional model. Our attentional model firstly extract features of depth image, and then the extracted features are used for both object classification (SC) and next-best-view prediction (3D-STN). Our 3D-STN will automatically regress the increment of current view. With current view and the view increment, our camera will move to next-best-view, and generate a new depth image. RNN  hidden  layers can help aggregate current depth image features with those from the past views. This is repeated until termination conditions are met.
We set two termination conditions for inference:
1) The classification uncertainty, measured by the Shannon entropy of the classification probability, is less than a threshold ($0.1$);
2) The maximum number ($10$) of time steps has been reached.

\section{Experimental setup}

\paragraph*{3D shape datasets.}
Our method is evaluated with three large-scale 3D shape datasets,
including ModelNet10~\cite{modelnet10}, ModelNet40~\cite{Wu15} and ShapeNetCore55~\cite{Su15}.
ModelNet10 and ModelNet40 are two subsets of Princeton ModelNet which contains 127,915 3D shapes and are categorized into 660 classes. ModelNet10 contains 10 categories with a total of 4899 3D shapes, and these shapes are split into train set and test set. Train set contains 3,991 shapes, while test set contains 908 shapes.  ModelNet40 is formed by 12,311 3D shapes which are categorized into 40 classes. All 3D shapes of ModelNet40 are also split into train set (9,843 3D shapes) and test set (2,468 3D shapes) respectively. ShapeNetCore55 is a  richly-annotated,  large-scale dataset  of  3D  shapes,  which has  55 common  object  categories  with  about  51,300  unique  3D shapes. For ShapeNetCore55, we split all shapes of each class into training (80\%) and testing (20\%) subsets.

Before training,  the center of 3D shape coincides with the origin of spherical coordinate system. And we scale the shape to the range of [-0.5, 0.5]. The radius of spherical coordinate system is set to be 1. During training,  a 3D shape needs to be rendered into a $227 \times 227$ depth image from any given viewpoint, using the ray casting algorithm (parallel implementation on the GPU). All depth values are normalized into $[0,255]$. If a ray does not hit the shape, the related depth value is set to 255.

\paragraph*{Parameters of subnetworks.}
We use the ReLU activation functions for all hidden layers, $f(x) = max(0,x)$, for fast training.
Shape classifier takes the AlexNet architecture~\cite{krizhevsky2012}
which contains $5$ convolutional layers and $3$ fully connected layers followed by a soft-max layer (Fig.~\ref{fig:2dstn}).
Same parameter settings as the original paper are used for the various layers.
The 3D-STN contains $5$ convolutional layers (shared with the shape classifier)
and $2$ fully connected layers.
In summary, there are about $62M$ parameters in total being optimized in training the 3D-STN, the RNN and the SC.
The maximum number of views for each view sequence are 10, the learning rates of pre-training SC is set to be 0.01, and the learning rates of joint training 3D-STN, SC and RNN is set to be 0.001.

\section{Results \& evaluation}
\label{sec:results}

\paragraph*{Hidden layer size.}
Obviously the size of hidden layer is an important hyper-parameter that affects the performance of recurrent neural networks. In this experiment, we evaluate the effect of hidden layer size. We carry out object recognition experiment on ModelNet40, and show the accuracy. The view number for each sequence are set to be 5, and the radius ($R$) remains 1.

The candidate sizes of hidden layer are $64$, $128$, $256$, $512$, and $1024$. All the results are show in Table~\ref{tab:hidden}. From the comparison of different hidden layer size, we note that our recurrent neural network achieves the best performance with a hidden layer size of $256$. All the following experiments are conducted with a fixed hidden layer size of $256$.
\begin{table}[!h]
\renewcommand\arraystretch{1.5}
\centering
\small
\begin{tabular}{l||c|c|c|c|c} \hline
                Hidden layer size    &  64     &  128    & 256     & 512   &  1024     
\\ \hline      Accuracy\ (\%)        &  87.1  &  87.4  & \textbf{88.3}   & 87.6 &  87.8    
\\ \hline
\end{tabular} 
\newline
\caption{Accuracy comparison under ModelNet40, $R$ remains 1, with the view number for each sequence being $5$.}
\label{tab:hidden}
\end{table}

\paragraph*{Effect of radius.}
We have set a fixed radius of spherical coordinate system in our previous experiments. However, the radius has an big impact on the recognition performance. If the radius is to large, the 3D shape will be a very small projection on depth image plane. The aim of this experiment is to find the best radius for our task.

All 3D shapes are scaled to the range of $[-0.5,0.5]$. We conduct a comparison over different radius: $0.5$, $1.0$, $1.5$, $2.0$, and $2.5$ over ModelNet40 with the view number being $5$. Results are shown in Table~\ref{tab:radius}.

\begin{table}[!h]
\renewcommand\arraystretch{1.5}
\centering
\small
\begin{tabular}{l||c|c|c|c|c} \hline
                Radius               &  0.5     &  1.0    & 1.5     & 2.0   &  2.5
\\ \hline      Accuracy\ (\%)        &  81.4    & \textbf{88.2}  & 82.6    & 75.8  &  69.3
\\ \hline
\end{tabular}
\newline
\caption{Accuracy comparison for different radius on ModelNet40, with the view number for each sequence being $5$.}
\label{tab:radius}
\end{table} 

As shown in Table~\ref{tab:radius}, our recurrent neural network achieves the best result with a radius of $1.0$, under the condition that we scale our 3D shape in a range of $[-0.5,0.5]$. Results show that camera should neither be too far nor too near to the 3D shape. On one hand, if the camera is too far to the 3D shape, the accuracy will drop dramatically. If the camera is too near to the 3D shape, on the other hand, the shape projection on depth image will be clipped, which will decrease the accuracy of recognition. Moreover, if the projection of 3D shape spans the full depth image, some details of shape border will be lost during the convolution operation.

\begin{figure}[t!] \centering
	\begin{overpic}[width=1.0\linewidth,tics=5]{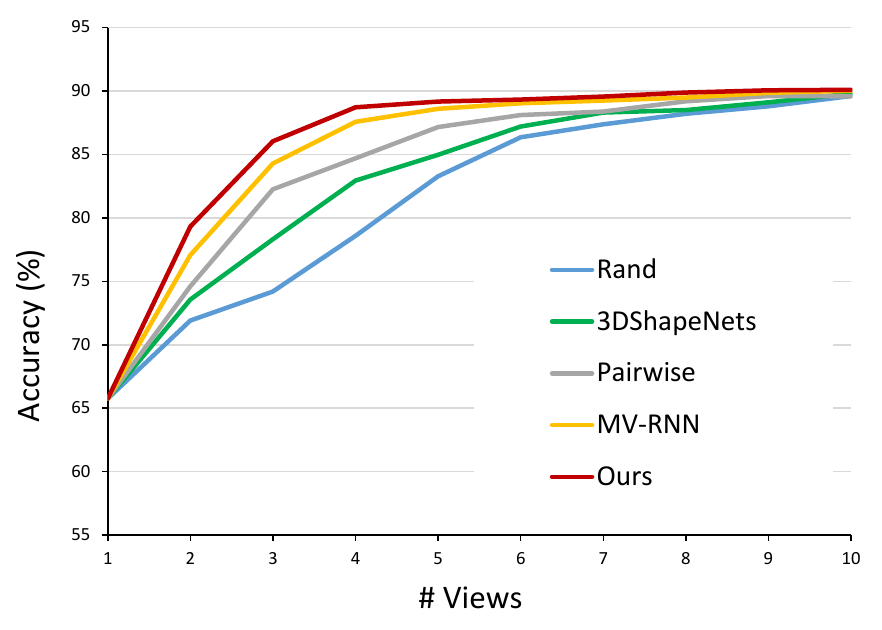}
	\end{overpic}
    \caption{Recognition accuracy over the number of views by five methods (\emph{Rand}, \emph{3DShapeNets}~\protect\cite{Wu15},
    \emph{MV-RNN}~\protect\cite{xu2016}, \emph{Pairwise}~\protect\cite{johns2016pairwise} and ours).}
    \label{fig:plot-nbv}
\end{figure}

\paragraph*{Comparison of NBV estimation.}
We tackle next-best-view (NBV) problem as seeking single camera views in order to improve accuracy of 3D shape recognition. Depending on this setting, NBV can be seen as an incremental approach to build up a sensing strategy for multi-view active object recognition, and NBV always tries to achieve the highest possible accuracy with the smallest number of views. But, with partial observation of a 3D shape, it is very hard to acquire the global optimal for NBV, so we set two criteria on evaluating NBV estimation: recognition accuracy and information gain.

To evaluate the performance of NBV estimation, we compare our attentional method against four alternatives,
including a baseline method and three state-of-the-art ones.
The baseline method \emph{Rand} selects the next view randomly.
The state-of-the-art NBV techniques being compared include
the \emph{3DShapeNets}~\cite{Wu15}, the \emph{MV-RNN}~\cite{xu2016} and the active recognition method based
on \emph{Pairwise} learning~\cite{johns2016pairwise}.
We train our model on the train set of ModelNet40 with the task of classification over the $40$ classes, and test on the test set of ModelNet40.
For a fair comparison on the quality of estimated NBVs, all method are evaluated only with depth images.
We let each method predict their own NBV sequences, and uniformly use a pre-trained Multi-View CNN shape classifier (MV-CNN)~\cite{Su15MV} for shape classification.
Fig.~\ref{fig:plot-nbv} plots the recognition rate over the number of predicted NBVs.
We note that our method achieves the fastest accuracy increase.

\begin{figure}[t!] \centering
	\begin{overpic}[width=1.0\linewidth,tics=5]{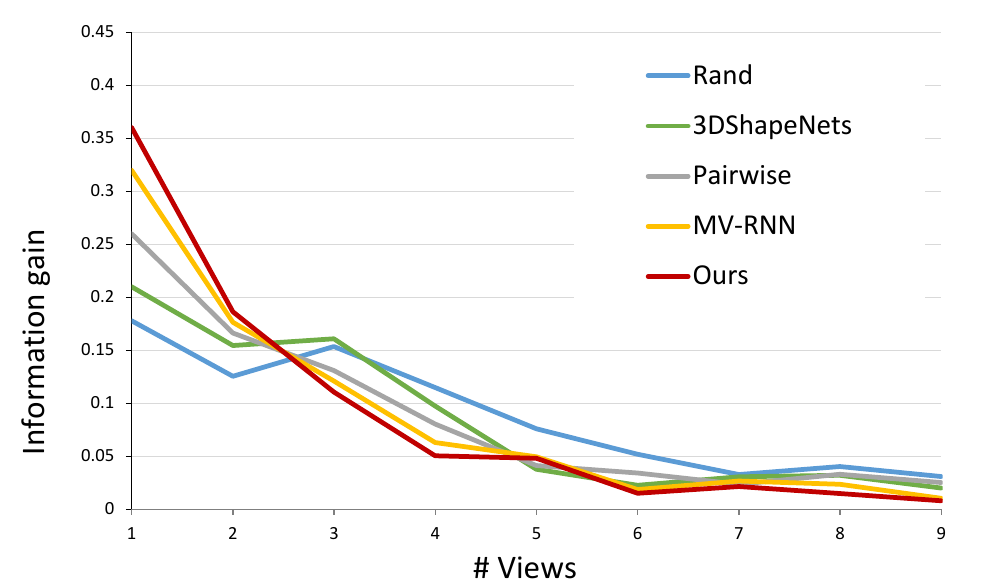}
	\end{overpic}
    \caption{Information gain of different views selection by five methods (\emph{Rand}, \emph{3DShapeNets}~\protect\cite{Wu15},
    \emph{MV-RNN}~\protect\cite{xu2016}, \emph{Pairwise}~\protect\cite{johns2016pairwise} and ours).}
    \label{fig:plot-inf}
\end{figure}

To further evaluate the efficiency of the NBVs, we plot in Fig.~\ref{fig:plot-inf} the information gain of different NBVs. The information gain is measured by the decrease of Shannon entropy of the classification probability distributions:
\begin{equation}
{I_t}({p_t},{p_{t - 1}}) = H({p_{t - 1}}) - H({p_t}),
\label{eq:inf}
\end{equation}
with $H(p) =  - \sum {_kp({y_k})} \log p({y_k})$. Compared to the alternatives, the NBV sequences output by our method attain larger information gain in early steps, demonstrating higher efficiency.



\begin{table}[!h]
\renewcommand\arraystretch{1.5}
\centering
\small
\begin{tabular}{l||c|c|c} \hline
 \multirow{2}{*}{Dataset}     & \multicolumn{2}{c|}{Train}        & \multirow{2}{*}{Test}      \\
 \cline{2-3}
 \multicolumn{1}{r||}{}       & Joint training    & Pre-training SC                                       \\
 \hline      ModelNet10    &  $7.0$ hr.      &  $2.7$ hr.      & \multirow{2}{*}{$0.1$ sec.}  \\
 \cline{1-3}
             ModelNet40    &  $14.3$ hr.     &  $5.2$ hr.      
\\ \hline
\end{tabular}
\newline
\caption{Training and testing time of our method.}
\label{tab:timing}
\end{table}

\begin{table}[!h]
\renewcommand\arraystretch{1.5}
\centering
\small
\begin{tabular}{l||c|c} \hline
                Method           &  Train        &  Test
 \\ \hline      3DShapeNets      &  $96$ hr.     &  $4$ min.
 \\ \hline      MV-RNN           &  $66$ hr.     &  $0.1$ sec.
 \\ \hline      Ours             &  $19.5$ hr.   &  $0.1$ sec.
\\ \hline
\end{tabular}
\newline
\caption{Comparing training and testing time between \emph{3DShapeNets}, \emph{MV-RNN} and our method.}
\label{tab:timecomp}
\end{table} 

\begin{table*}[!ht]
  \renewcommand\arraystretch{1.5}
  \centering
  \fontsize{7.5}{8}\selectfont

  \begin{tabular}{c||c|c|c|c|c|c|c|c|c|c|c|c|c}
   \hline

   \multirow{2}{*}{Method} &\multirow{2}{*}{Image} &\multicolumn{4}{|c|}{ModelNet10}&\multicolumn{4}{|c|}{ModelNet40}&\multicolumn{4}{c}{ShapeNetCore55}\cr\cline{3-14}
   & &3 views&6 views & 9 views&Average &3 views &6 views &9 views &Average&3 views & 6 views& 9 views&Average\cr\hline
   3DShapeNets & Depth&76.9 &81.8 &82.5 &80.4 &71.2 &74.8 &78.1 &74.7 &69.3 &71.7 &72.8 &71.3  \cr\hline
    MV-RNN     & Depth&86.4 &88.9 &89.5 &88.2 &84.3 &86.5 &88.6 &86.4 &71.1 &73.8 &76.3 &73.7 \cr\hline
    Pairwise   & Depth&84.9 &87.7 &89.2 &87.3 &82.7 &85.1 &88.3 &85.3 &70.8 &73.9 &76.2 &73.6 \cr\hline
    Ours       & Depth&87.7 &89.8 &91.2 &\textbf{89.5} &86.1 &88.7 &89.8 &\textbf{88.2} &71.6 &74.5 &76.9 &\textbf{74.3} \cr\hline

  \end{tabular}
  \newline
   \caption{Recognition accuracy comparison on four methods with three datasets.}
  \label{tab:classification}
\end{table*}

\paragraph*{Comparison of 3D shape recognition.}
To  evaluate  the performance of recognition, we carry out comparison experiment on three datasets: ModelNet10, ModelNet40 and ShapeNetCore55. We compare our method against recent competing methods: \emph{3DShapeNets}~\cite{Wu15}, \emph{MV-RNN}~\cite{xu2016} and \emph{Pairwise} learning~\cite{johns2016pairwise}. For a fair comparison, all method are evaluated only with depth images. For \emph{3DShapeNets} and \emph{Pairwise} learning, we select the next view along the sequence by following a straight path around the viewing sphere from the beginning to the end of the sequence~\cite{johns2016pairwise}.

Table~\ref{tab:classification} shows the recognition results for a random initial view from our selected 50 views. Our method achieves the best results, and we note that \emph{3DShapeNets}~\cite{Wu15}, \emph{MV-RNN}~\cite{xu2016} and our methods are depth-based, while \emph{Pairwise} learning~\cite{johns2016pairwise} uses both greyscale and depth images. In our experiment, however, we only use deep images for \emph{Pairwise} learning for a fair comparison.

We also notice that VERAM~\cite{veram2018} achieves recognition accuracy of 92.1\% on ModelNet40 with 9 views of gray images. They align all shapes and render 144 gray images for each shape with Phong reflection model. Reinforcement learning is adopted to solve the problem that gradient from observation subnetwork can not be back propagated to recurrent subnetwork. They integrate view confidence and view location constrains into reward function. Moreover, three strategies (Sign, Clamp and ELU) are deployed to enhance gradient. For RNN, they deploy Long Short-Term Memory (LSTM) units. For a fair comparison, we test with three experimental settings of VERAM: First, we change 144 viewpoints to 50 viewpoints; Second, we use our ray casting method to generate depth images instead Phong gray images; Third, we modify LSTM of RNN with linear mapping to keep the same setting as our method. Both VERAM and our method use AlexNet and the same radius. We find that, under the setting of modified VERAM, the recognition accuracy of ModelNet40 with 9 views is 88.7\%, which is inferior to our method (89.8\%).

\begin{figure*}[t!] \centering
	\begin{overpic}[width=1.0\linewidth,tics=5]{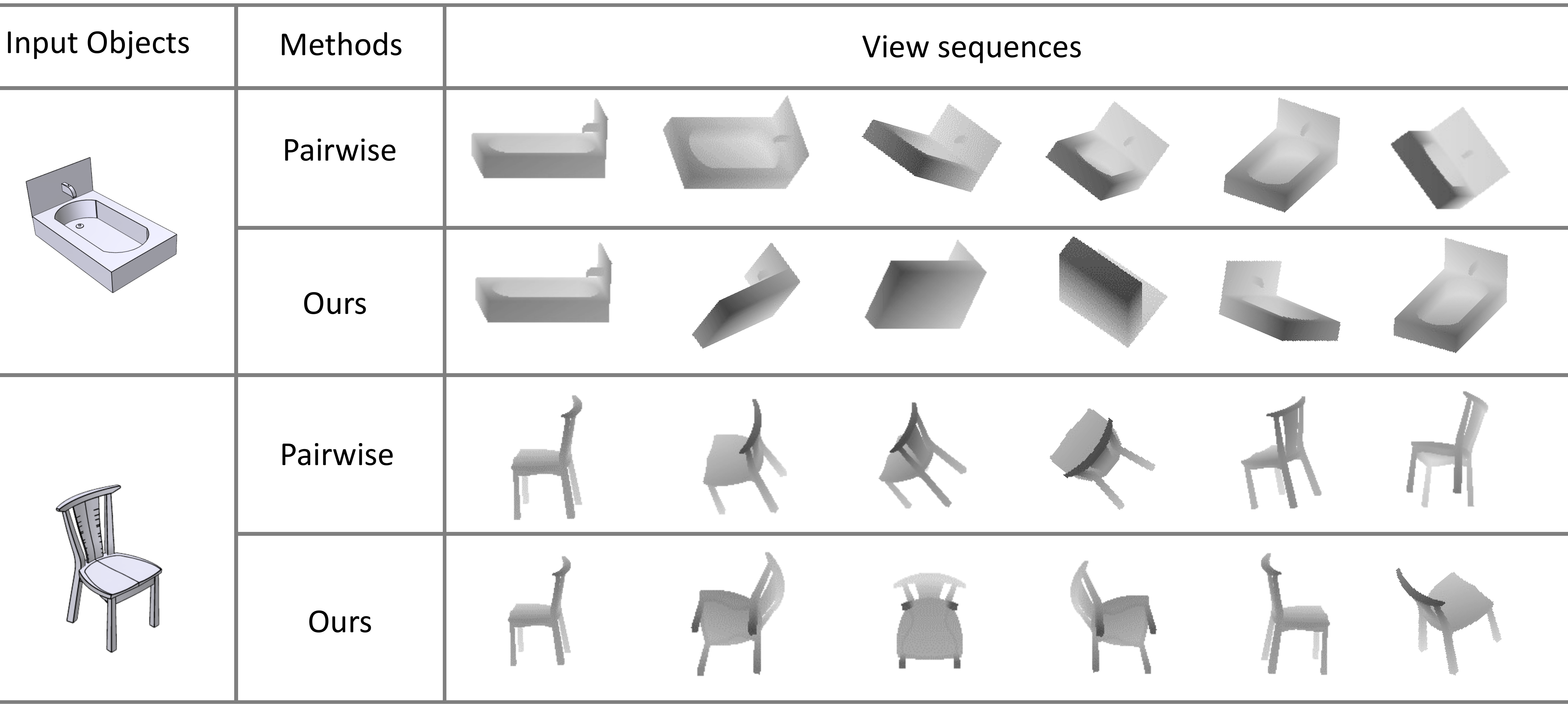}
	\end{overpic}
    \caption{Comparison of view sequences on \emph{Pairwise} learning~\cite{johns2016pairwise} and ours (input objects are from ModelNet40).}
    \label{fig:vis_comp}
\end{figure*}
\begin{figure*}[t!] \centering
	\begin{overpic}[width=1.0\linewidth,tics=5]{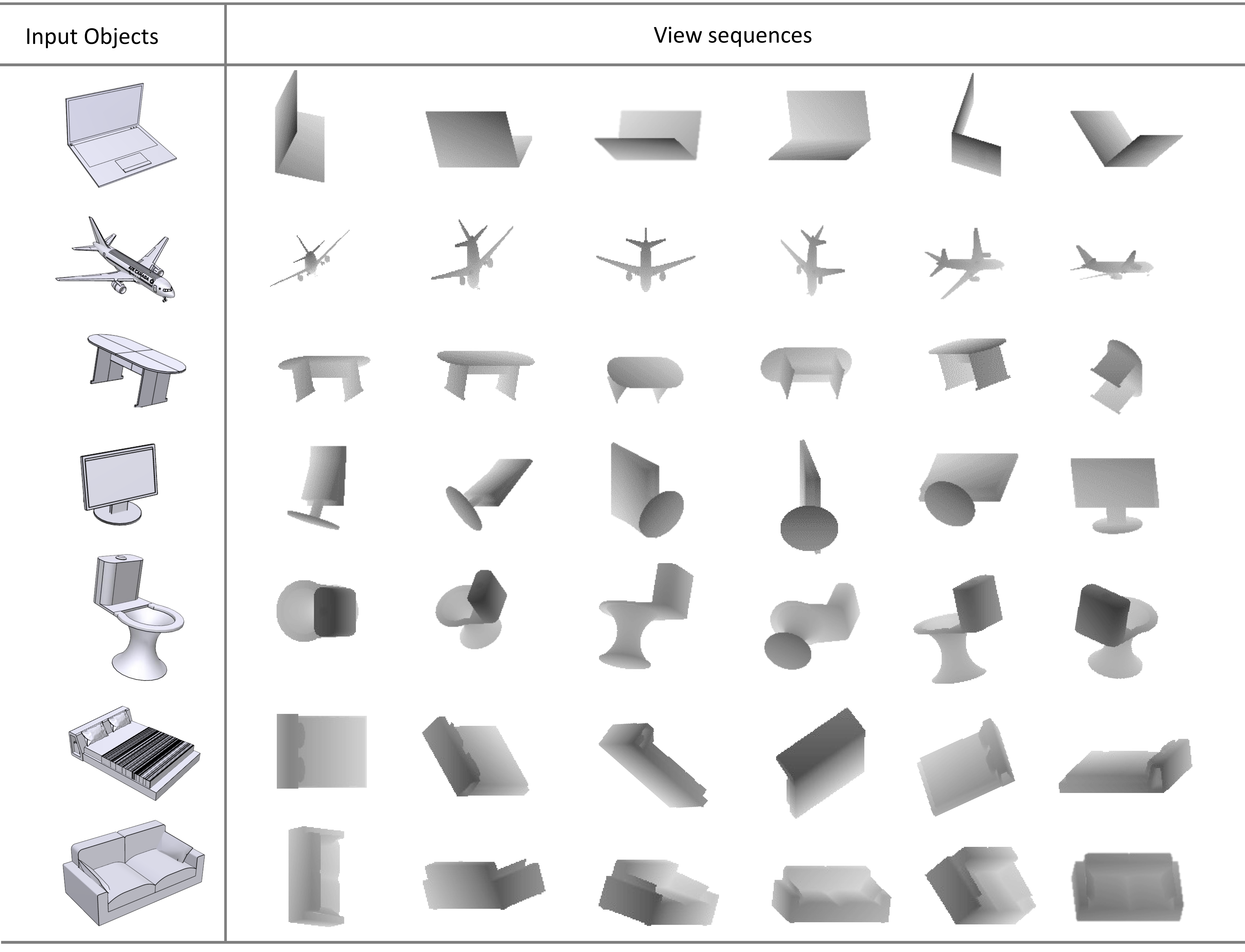}
	\end{overpic}
    \caption{Visualization of the active recognition process by showing the NBV sequence with depth images (input objects are from ModelNet40).}
    \label{fig:vis}
\end{figure*}

\paragraph*{Timings.}
Table~\ref{tab:timing} lists the training and testing time of our method
on both ModelNet10 and ModelNet40 datasets. Since the shape classifier is pre-trained outside the joint training networks (3D-STN, RNN and SC), we report its pre-train time separately.

Table~\ref{tab:timecomp} compares the training time of three methods, i.e.,
\emph{3DShapeNets}, \emph{MV-RNN} and ours on ModelNet40.
The training of \emph{3DShapeNets} involves learning the generative model with CDBN.
\emph{MV-RNN} is trained with reinforcement learning.
The comparison shows the training efficiency of our model over the two alternatives.
All timings were obtained on a workstation with an Intel$^{\circledR}$ Xeon E5-2670 @ 2.30GHz $\times$ 24, 64GB RAM
and an Nvidia$^{\circledR}$ Titan Xp graphics card with 12GB memory.



%


\paragraph*{Visualization.}
To visually investigate the behavior of our model,
we visualize in Fig.~\ref{fig:vis_comp} for view sequences comparison between \emph{Pairwise} learning~\cite{johns2016pairwise} and our method. The results demonstrate that our method can correctly recognize the objects with plausibly planned views. We note that the regressed view sequences tends to have a whole coverage of shapes for higher recognition rates than \emph{Pairwise} learning. In our method, we start training our model from separate 50 initial views, which means we can start from different initial views when testing. More results of our method are shown in Fig.~\ref{fig:vis}. All input objects are from the test set of ModelNet40.


\section{Conclusion}
\label{sec:conclusion}

We have proposed a 3D attentional formulation to the active object recognition problem.
This is mainly motivated by the mechanism resemblance between human attention and active
perception~\cite{bajcsy1988active}, and the significant progress made in utilizing attentional models
to address complicated vision tasks such as image captioning~\cite{Xu2015Cap}.
In developing such a model, we utilize RNNs for learning and storing the internal representation of the object being observed, CNNs for performing depth-based recognition, and STNs for selecting the next-best-views. The carefully designed 3D STN makes the whole network differentiable and hence easy to train. Experiments on well-known datasets demonstrate the efficiency and robustness of our active recognition model.

\paragraph*{Limitations.}
A drawback of learning a policy offline is that physical restrictions during testing are
hard to incorporate and when the environment is changed, the computed policy would no longer be useful.
This problem can be alleviated by learning from a large amount of training data and cases using
a high capacity learning model such as deep neural networks as we do.
Our method does not handle mutual occlusion between objects which is a frequent case in cluttered scenes.
One possible solution is to train the recognition network using depth images with synthetic occlusion.

\paragraph*{Future works.}
In the future, we would like to investigate a principled solution for handling object occlusion in real indoor scenes,
e.g., using STN to help localize the shape parts which are both visible and discriminative, in a similar
spirit to~\cite{xiao2015}. Another interesting direction is to study multi-agent attentions
in achieving cooperative vision tasks, such as multi-robot scene reconstruction and understanding.
It is particularly interesting to study the shared and distinct attentional patterns
among heterogeneous robots such as mobile robots and drones.


\CvmAck{We thank the anonymous reviewers for their valuable comments. This work was supported, in parts, by an NSFC programs (61572507, 61622212, 61532003). Min Liu is supported by the China Scholarship Council.}

\bibliographystyle{CVM}

{\normalsize  \bibliography{ref}}

\Author{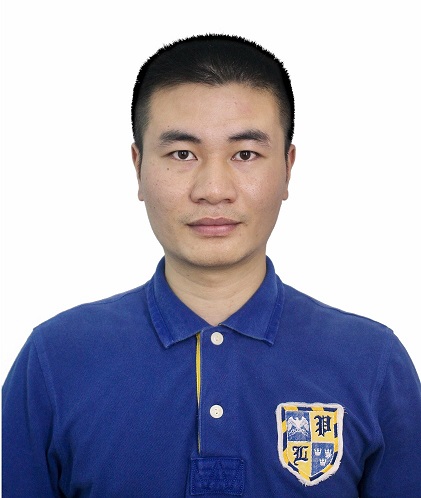}{Min Liu}
{is a Ph.D. candidate in School of Computer, National University of Defense Technology. He received his BS degree in geodesy and geomatics from Wuhan University and MS degree in computer science from National University of Defense Technology in 2013 and 2016, respectively. He is visiting University of Maryland at College Park from 2018 to 2020. His research interests mainly include robot manipulation and 3D vision.}

\Author{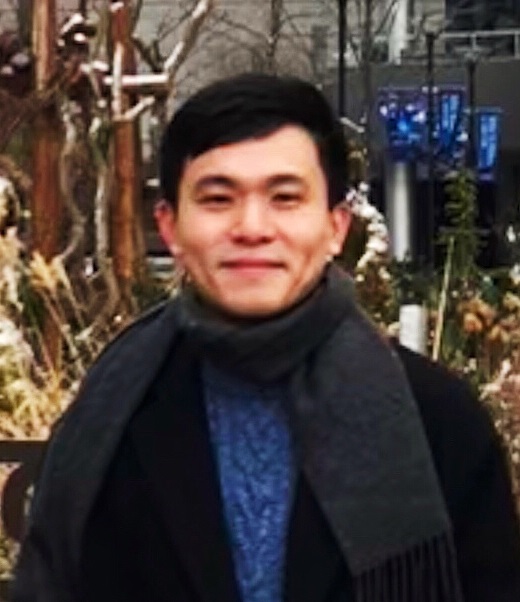}{Yifei Shi}
{received his BS degree in geodesy and geomatics from Wuhan University and MS degree in computer science from National University of Defense Technology in 2012 and 2015, respectively. He is pursing a doctorate in computer science at National University of Defense Technology. His research interests mainly include data-driven scene understanding, RGBD reconstruction and 3D vision.}

\Author{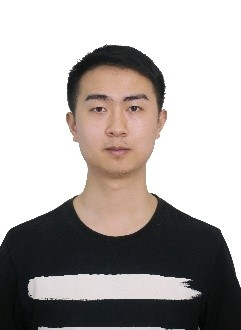}{Lintao Zheng}
{received his BS degree in applied
mathematics from Xian Jiaotong University and MS degree in computer science from the National University of Defense Technology in 2013 and 2016, respectively. He is pursuing a doctorate in computer science at the National University of Defense Technology. His research interests mainly include computer graphics, deep learning and robot vision.}

\Author{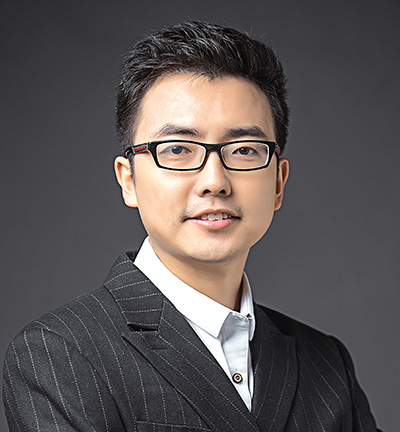}{Kai Xu}
{is an Associate Professor at the School of Computer, National University of Defense Technology, where he received his Ph.D. in 2011. He conducted visiting research at Simon Fraser University during 2008-2010, and Princeton University during 2017-2018. His research interests include geometry processing and geometric modeling, especially on data-driven approaches to the problems in those directions, as
well as 3D-geometry-based computer vision. He has published over 60 research papers, including 21 SIGGRAPH/TOG papers. He organized two SIGGRAPH Asia courses and one Eurographics STAR tutorial. He is currently serving on the editorial board of Computer Graphics Forum, Computers \& Graphics, and The Visual Computer. He also served as paper co-chair of CAD/Graphics 2017 and ICVRV 2017, as well as PC member for several prestigious conferences including SIGGRAPH Asia, SGP, PG, GMP, etc. His research work can be found in his personal website:www.kevinkaixu.net.}

\Author{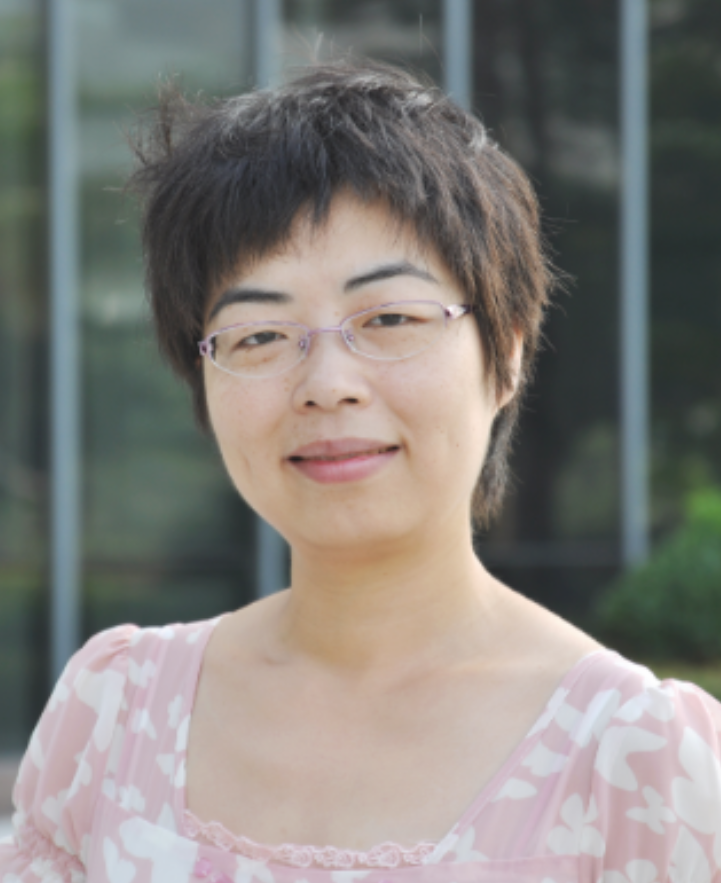}{Hui Huang}
{is Distinguished Professor, Founding Director of the Visual Computing Research Center, Shenzhen University. She received her PhD degree in Applied Math from the University of British Columbia (UBC) in 2008 and another PhD degree in Computational Math from Wuhan University (WHU) in 2006. Her research interests are in Computer Graphics and Vision, focusing on Geometric Modeling, Shape Analysis, Point Optimization, Image Processing, 3D/4D Acquisition and Creation. She is currently an Associate Editor-in-Chief of The Visual Computer (TVC) and is on the editorial board of Computers \& Graphics (CAG) and Frontiers of Computer Science (FCS). She has served on the program committees of almost all major computer graphics conferences including SIGGRAPH ASIA, EG, SGP, PG, 3DV, CGI, GMP, SMI, GI and CAD/Graphics. She is invited to be CHINAGRAPH 2018 Program Vice-Chair, in addition to SIGGRAPH ASIA 2017 Technical Briefs and Posters Co-Chair, SIGGRAPH ASIA 2016 Workshops Chair and SIGGRAPH ASIA 2014 Community Liaison Chair. She is the recipient of NSFC Excellent Young Scientist, Guangdong Technological Innovation Leading Talent Award, CAS Youth Innovation Promotion Association Excellent Member Award, Guangdong Outstanding Graduate Advisor Award, CAS International Cooperation Award for Young Scientists, CAS Lujiaxi Young Talent Award. She is also selected as CCF Distinguished Member and ACM/IEEE Senior Member.}

\Author{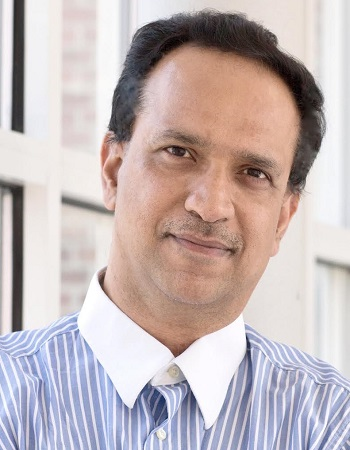}{Dinesh Manocha}
{is the Paul Chrisman Iribe Chair in Computer Science \& Electrical and Computer Engineering at the University of Maryland at College Park. He is also the Phi Delta Theta/Matthew Mason Distinguished Professor Emeritus of Computer Science at the University of North Carolina - Chapel Hill. He has won many awards, including Alfred P. Sloan Research Fellow, the NSF Career Award, the ONR Young Investigator Award, and the Hettleman Prize for scholarly achievement. His research interests include multi-agent simulation, virtual environments, physically-based modeling, and robotics. His group has developed a number of packages for multi-agent simulation, crowd simulation, and physics-based simulation that have been used by hundreds of thousands of users and licensed to more than 60 commercial vendors. He has published more than 480 papers and supervised more than 35 PhD dissertations. He is an inventor of 9 patents, several of which have been licensed to industry. His work has been covered by the New York Times, NPR, Boston Globe, Washington Post, ZDNet, as well as DARPA Legacy Press Release. He is a Fellow of AAAI, AAAS, ACM, and IEEE and also received the Distinguished Alumni Award from IIT Delhi. See http://www.cs.umd.edu/dm.}

\end{document}